\documentclass{llncs}
\usepackage[T1]{fontenc}
\usepackage{graphicx}
\usepackage{amsmath,amssymb,amsfonts}
\usepackage{booktabs}
\usepackage{array}
\usepackage{microtype}
\usepackage{float}
\usepackage{hyperref}
\hypersetup{colorlinks=true, linkcolor=blue, citecolor=blue, urlcolor=blue}
\begin{document}
\pagestyle{plain}
\title{Knowledge Distillation from Large Reasoning Models to Compact Student Models: A Case Study on the John O'Bryan Mathematics Competition}
\author{Gaurab Baral\inst{}\thanks{Corresponding author: \email{baralg1@nku.edu}} \and
        Aaditya Khanal\inst{} \and
        Yangyang Tao\inst{} \and
        Junxiu Zhou\inst{}}
\institute{Northern Kentucky University,
Highland Heights, KY 41099, USA\\
\email{\{baralg1, khanala1, taoy1, zhouj2\}@nku.edu}}
\maketitle

\begin{abstract}
This paper investigates knowledge distillation from a large reasoning model (DeepSeek-R1) to a compact student model (Qwen2.5-7B). Using historical problems from the John O'Bryan Mathematics Competition at Northern Kentucky University (2011--2025), we build a Chain-of-Thought (CoT) training corpus through a dual-agent framework. The dataset is used to fine-tune the student model with Low-Rank Adaptation (LoRA) on Apple Silicon hardware using the MLX framework~\cite{mlx2023}. The base Qwen2.5-7B model achieves 64.67\% accuracy on competition problems, while the DeepSeek-R1 teacher achieves 91.40\%. An initial 1,000-iteration training run revealed severe overfitting, with validation loss reaching a minimum at iteration~200 before rising steadily. Based on this finding, we ran five independent training runs each limited to 200 iterations with varied random seeds to assess result stability. Across these five runs, the fine-tuned student model achieves a mean accuracy of 69.43\% $\pm$ 0.17\% on the competition dataset, a 4.76 percentage-point improvement over the base model, and generalizes to 73.1\% $\pm$ 0.18\% on the MATH-500 benchmark. We further study how response length affects answer quality across six reasoning levels (R1--R6): accuracy declines consistently from 69.43\% at R1 (mean 220 words) to 41.9\% at R6 (mean 31.2 words), with the two-person speed section most sensitive to token reduction. These results demonstrate that CoT distillation improves compact student models and that response length is a critical factor in mathematical reasoning quality.
\end{abstract}


\clearpage

%-------------------------------------------------------------------
\section{Introduction}
%-------------------------------------------------------------------

Large language models (LLMs) capable of advanced mathematical reasoning have become widely used in recent years. Models such as DeepSeek-R1 reach strong performance on competition-level mathematics benchmarks, but their size makes local deployment impractical. Knowledge distillation offers a way to address this: by transferring the reasoning behavior of a large teacher model into a smaller student model, one can produce compact models that approach the teacher's performance at substantially lower computational cost~\cite{gou2021}.

Despite recent progress in distilling reasoning ability from large models~\cite{deepseek2025,hinton2015}, most studies rely on generic benchmarks such as MATH-500~\cite{hendrycks2021} or GSM8K~\cite{cobbe2021}, which may include data seen during pretraining. There is a gap in understanding how distillation performs on domain-specific, expert-curated problem sets with lower contamination risk.

This work addresses that gap by applying knowledge distillation to undergraduate mathematics competition problems using the John O'Bryan Mathematics Competition at Northern Kentucky University (NKU) as our problem corpus. The competition covers multiple difficulty levels and mathematical topics, making it a focused and well-defined dataset for evaluating LLM reasoning. To our knowledge, this is the first study to apply CoT distillation to this corpus and to quantify the effect of token budget constraints across six reasoning levels on competition-style problems.

Our main contributions are as follows: (1) we construct a CoT training corpus from 15 years of competition problems using a dual-agent teacher-verifier pipeline; (2) we demonstrate that LoRA fine-tuning with early stopping improves Qwen2.5-7B from 64.67\% to 69.43\% $\pm$ 0.17\%, generalizing to MATH-500; (3) we quantify how accuracy degrades as token budgets are reduced across six levels, identifying a practical lower bound of roughly 50--100 words for multi-step competition problems; and (4) we provide an error-type analysis showing that roughly 40\% of failures are formatting rather than mathematical errors.

The pipeline proceeds as follows. First, we collect and digitize competition problems from 2011 to 2025. Second, a dual-agent framework powered by the DeepSeek-R1 API produces CoT reasoning traces and verifies answers. Third, a quantized Qwen2.5-7B student model is fine-tuned with LoRA using the MLX framework. A diagnostic 1,000-iteration run identifies the best early stopping point, after which five independent 200-iteration runs with different random seeds produce stable results. Finally, we evaluate the fine-tuned models on both the competition dataset and MATH-500, and run a multi-level token budget experiment. The full implementation, including data preparation scripts, fine-tuning code, evaluation pipelines, and raw result files, is publicly available at \url{https://github.com/TempGaurab/Distillation.John-O-Bryan}.

%-------------------------------------------------------------------
\section{Related Work}
%-------------------------------------------------------------------

\subsection{Knowledge Distillation}

Knowledge distillation was introduced by Hinton, Vinyals, and Dean~\cite{hinton2015}, who showed that a smaller network can approximate a larger ensemble by training on soft output targets. Soft targets carry richer class-relationship information than one-hot labels, helping the student generalize beyond the training distribution. Gou et al.~\cite{gou2021} survey the extensive literature that followed, documenting how distillation techniques have expanded to cover feature-level, relation-based, and task-specific objectives across many model types. Boix-Adsera~\cite{boix2024} provides PAC-theoretic justification, proving that student models trained from a teacher can achieve lower sample complexity than learning from raw labels alone.

\subsection{Distillation for Reasoning LLMs}

Applying distillation to reasoning-focused LLMs has gained traction since 2023. DeepSeek-R1~\cite{deepseek2025} reached state-of-the-art performance on MATH-500 and AIME benchmarks using reinforcement learning, and its technical report demonstrates that distilled smaller variants (e.g., DeepSeek-R1-Distill-Qwen-32B) retain much of the teacher's reasoning ability. Deng et al.~\cite{deng2023} propose implicit chain-of-thought distillation, where teacher reasoning is encoded into student hidden states rather than surface-level text, improving GSM8K performance without generating explicit intermediate steps. Ho et al.~\cite{ho2022} show that fine-tuning smaller GPT-2 variants on teacher-generated CoT traces substantially improves multi-step arithmetic accuracy, providing early evidence that reasoning traces are effective training signal. Magister et al.~\cite{magister2023} extend this finding to T5 models of varying sizes, showing that even 11B-parameter students benefit from distilled CoT supervision on grade-school and symbolic reasoning tasks. Parameter-efficient fine-tuning methods such as LoRA~\cite{hu2022} and prefix tuning have been shown to be effective for adapting LLMs to specific domains with limited data~\cite{zhang2023,ding2023}.

\subsection{Vision-Language Distillation}

Knowledge distillation has also been extended to the vision-language domain. The Vision-Language-Vision (VLV) auto-encoder framework~\cite{zhang2026} introduces a cost-efficient approach to building strong image-captioning models by using a frozen text-to-image diffusion model as an information bottleneck, then fine-tuning an LLM to decode the resulting language representations into detailed captions. The method achieves performance comparable to GPT-4o and Gemini 2.0 Flash at under \$1,000 training cost, demonstrating that strategic reuse of pretrained components can dramatically reduce data and compute requirements. This principle of leveraging a frozen teacher to generate high-quality training signal without retraining parallels our use of DeepSeek-R1 as a fixed API-based teacher.

\subsection{Response Length and Reasoning Quality}

The relationship between response length and model accuracy has received growing attention. Wei et al.~\cite{wei2022} show that chain-of-thought prompting substantially improves performance on multi-step arithmetic and commonsense problems, establishing that explicit intermediate steps are beneficial rather than redundant. Inference-time compute scaling results~\cite{deepseek2025} further confirm that allowing models more tokens to reason improves downstream accuracy on hard benchmarks. Related work on structuring intermediate reasoning steps, such as the subgoal-based demonstration learning framework of Zhao et al.~\cite{hou2024} for formal theorem proving, similarly suggests that the organization and granularity of intermediate reasoning steps materially affects downstream task accuracy, motivating closer study of how compressing those steps under a token budget affects performance. Our token budget experiment directly measures this tradeoff across six reasoning levels on a controlled competition problem set.

\subsection{Mathematical Benchmarks and Competition Problems}

Standard benchmarks such as MATH-500~\cite{hendrycks2021} and GSM8K~\cite{cobbe2021} are widely used for evaluating mathematical reasoning. Lightman et al.~\cite{lightman2023} show that process reward models trained on human-annotated solution steps outperform outcome-supervised models on MATH, highlighting the importance of step-level supervision. Competition-level problem sets offer expert curation, graded difficulty, and reduced pretraining contamination risk. The John O'Bryan Competition provides such a corpus in a controlled academic setting, spanning algebra, calculus, combinatorics, and linear algebra across three difficulty tiers.

%-------------------------------------------------------------------
\section{System Model and Methodology}
%-------------------------------------------------------------------

\subsection{Theoretical Foundation}

Knowledge distillation trains a student model $f_S$ to approximate a teacher $f_T$. The standard objective is:
\begin{equation}
  \mathcal{L} = (1-\alpha)\,\mathcal{L}_{CE}\!\left(y,\,f_S(x)\right)
              + \alpha\,\mathcal{L}_{KL}\!\left(f_T(x),\,f_S(x)\right)
  \label{eq:distill}
\end{equation}
where $\mathcal{L}_{CE}$ is the cross-entropy loss against ground-truth labels $y$, $\mathcal{L}_{KL}$ is the KL divergence between teacher and student output distributions, and $\alpha \in [0,1]$ balances the two terms. Because DeepSeek-R1 is accessed via API and full logit distributions are unavailable, we adopt a CoT distillation approach: the teacher generates step-by-step reasoning traces that serve as supervised training targets for the student. This reduces Eq.~(\ref{eq:distill}) to standard cross-entropy fine-tuning on the teacher's output sequences~\cite{ho2022}.

\subsection{Dataset Construction}

The corpus consists of 671 problems from the John O'Bryan Mathematics Competition (2011--2025), extracted from official PDF exam booklets using a combination of text extraction and manual digitization to ensure OCR quality. Problems with missing answer keys, illegible diagrams, or ambiguous phrasing were removed. The dataset was split chronologically by year: problems from 2011--2021 form the training set, 2022--2023 form the validation set, and 2024--2025 form the held-out test set. This year-based split prevents data leakage between phases. Records are stored as JSONL with fields: \texttt{year}, \texttt{section}, \texttt{question\_id}, \texttt{problem}, and \texttt{answer}, and formatted for the MLX LoRA framework.

\subsection{Dual-Agent Teacher Framework}

Two DeepSeek-R1 agents operate sequentially: \textbf{Agent 1 (Solver)} produces a step-by-step CoT response ending with a clearly delimited final answer; \textbf{Agent 2 (Verifier)} receives the original problem, the ground-truth answer, and Agent 1's response, then returns a \texttt{CORRECT}/\texttt{INCORRECT} verdict with brief justification. On the competition dataset (651 questions), the pipeline reached 91.4\% accuracy (594 correct); on MATH-500, 94.0\% (470/500). Only verified-correct traces were retained as training examples. While using DeepSeek-R1 as both teacher and verifier introduces a dependency, the verifier checks factual correctness against an independently provided ground truth, which partially mitigates stylistic bias.

\subsection{LoRA Fine-Tuning Configuration}

The student is Qwen2.5-7B-Instruct in 4-bit quantized form. LoRA~\cite{hu2022} adds rank-decomposition matrices $\Delta W = BA$ ($B \in \mathbb{R}^{d \times r}$, $A \in \mathbb{R}^{r \times k}$, $r \ll \min(d,k)$) while freezing the base weights. Configuration: rank $r=8$; $\alpha=16$; dropout $=0.05$; target modules \texttt{q\_proj}, \texttt{k\_proj}, \texttt{v\_proj}, \texttt{o\_proj}; context length 2,048 tokens; 4-bit (Q4) quantization. Trainable parameters: 11.534M (0.151\% of 7,615.617M), keeping the parameter count small to limit overfitting on a compact corpus~\cite{ding2023}.

\subsection{Multi-Run Training Protocol}

\textbf{Phase 1 (Diagnostic run):} We ran \texttt{mlx\_lm.lora} for 1,000 iterations with batch size 4, learning rate $1\times10^{-5}$, and gradient checkpointing enabled. Validation loss reached a minimum of 0.374 at iteration 200, then rose to 0.826 by iteration 1,000, while training loss fell toward zero. This identified iteration 200 as the optimal early stopping point.

\textbf{Phase 2 (Five independent runs):} Five training runs, each limited to 200 iterations, used different random seeds (42, 123, 256, 512, 999) for weight initialization and data shuffling, with all other hyperparameters held constant. Adapter weights from each run were merged into the base model using \texttt{mlx\_lm.fuse}, and all five merged models were evaluated separately. Peak memory per run: 12.247 GB.

\subsection{Multi-Level Token Budget Experiment}

We evaluated the fine-tuned model at six reasoning levels (R1--R6). R1 is the unrestricted baseline, corresponding to the 69.43\% fine-tuning result. R2 through R6 apply progressively smaller maximum token budgets: R2 (400 tokens), R3 (320 tokens), R4 (180 tokens), R5 (135 tokens), and R6 (100 tokens). Correctness was judged by the same DeepSeek-R1 judge for all levels. Only questions with a verified correct ground-truth answer (\texttt{correct\_flag == 1}) were included in accuracy calculations.

%-------------------------------------------------------------------
\section{Dataset Description}
%-------------------------------------------------------------------

The John O'Bryan Mathematics Competition, held annually at NKU, provided 671 problems across 15 years (2011--2025). After filtering, the corpus spans three sections: Freshman/Sophomore (265 problems: algebra, number theory, combinatorics, pre-calculus), Junior/Senior (268 problems: calculus, linear algebra, abstract reasoning), and Two-Person Speed (118 problems: fast multi-step numerical reasoning). A chronological train/validation/test split (2011--2021 / 2022--2023 / 2024--2025) prevents temporal leakage. Each record is a JSONL object with fields \texttt{year}, \texttt{section}, \texttt{question\_id}, \texttt{problem}, and \texttt{answer}; only the 594 teacher-verified correct traces were used for training. MATH-500~\cite{hendrycks2021} (500 problems, 7 subjects) served as out-of-domain evaluation; no MATH-500 problems appeared in training. Table~\ref{tab:dataset-stats} summarizes corpus statistics.

\begin{table}[H]
  \caption{John O'Bryan Competition Dataset Statistics}
  \label{tab:dataset-stats}
  \centering
  \begin{tabular}{lc}
    \toprule
    \textbf{Attribute} & \textbf{Value} \\
    \midrule
    Years / total problems            & 2011--2025 (15 yrs) / 671 \\
    Teacher-attempted / verified      & 651 / 594 \\
    F/S / J/S / Speed problems        & 265 / 268 / 118 \\
    Train / Val / Test split          & 2011--2021 / 2022--23 / 2024--25 \\
    Fine-tuning runs (seeds)          & 5 (42, 123, 256, 512, 999) \\
    Out-of-domain eval                & MATH-500 (500 problems, 7 subjects) \\
    \bottomrule
  \end{tabular}
\end{table}

\begin{figure}[H]
  \centering
  \includegraphics[width=\textwidth]{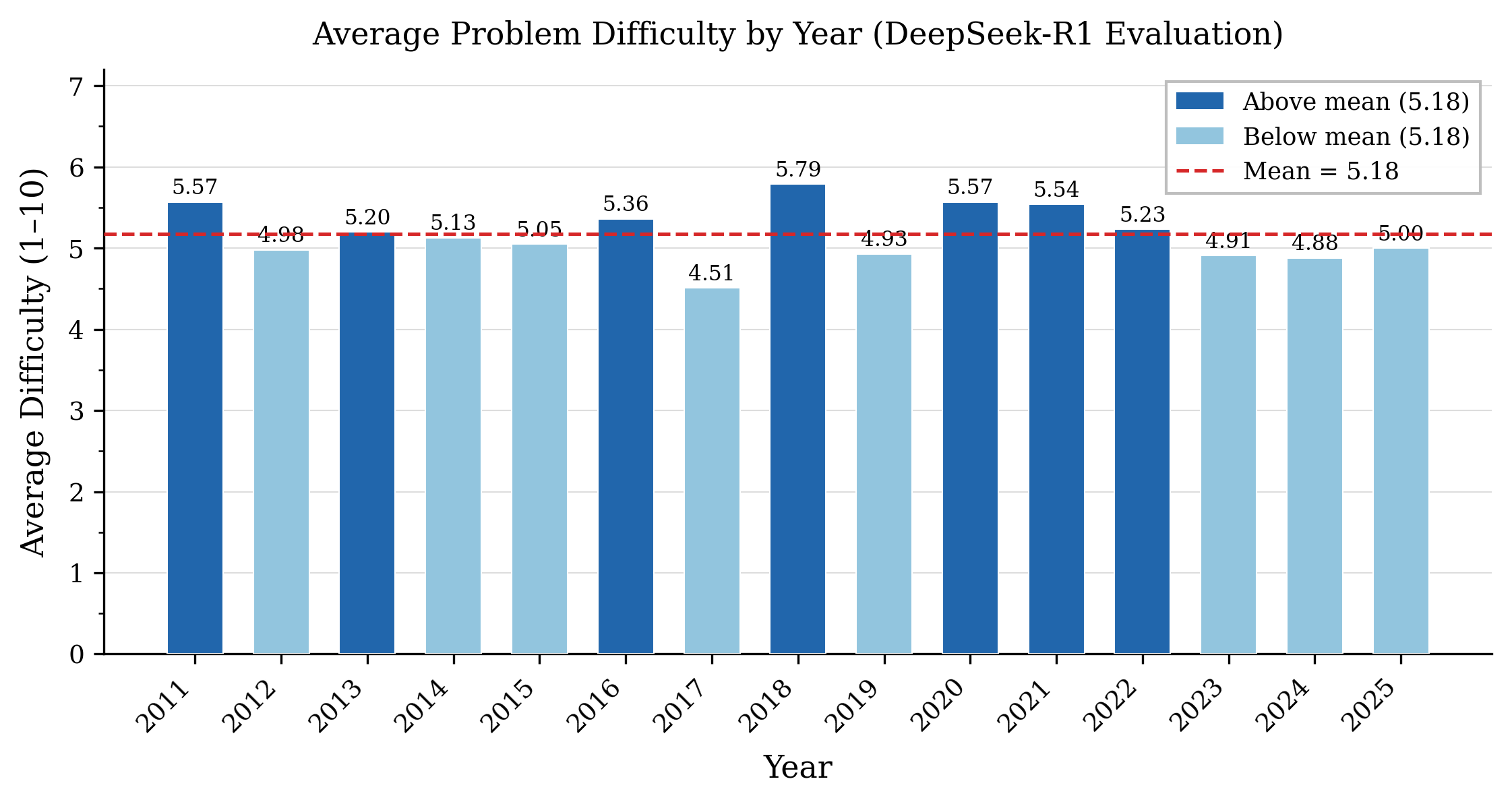}
  \caption{Average problem difficulty by year (DeepSeek-R1 ratings, scale 1--10). Darker bars are at or above the mean (5.19).}
  \label{fig:difficulty}
\end{figure}

DeepSeek-R1 rated each problem's difficulty on a 1--10 scale. Fig.~\ref{fig:difficulty} shows per-year averages ranging from 4.51 (2017) to 5.79 (2018), indicating moderate and consistent difficulty. Year 2017, the easiest by this metric, is where the fine-tuned model performs best across all reasoning levels.

%-------------------------------------------------------------------
\section{Evaluation Metrics}
%-------------------------------------------------------------------

Answer accuracy is the share of problems where the model's answer matches the ground truth:
\begin{equation}
  \text{Accuracy} = \frac{\text{Number of Correct Answers}}{\text{Total Problems Judged}}
\end{equation}
Because mathematical answers can appear in multiple equivalent forms, correctness is assessed by a separate DeepSeek-R1 judge rather than string matching. All task-level metrics from the main distillation experiment are reported as mean $\pm$ standard deviation across the five runs. Test perplexity is computed as $\text{PPL} = \exp(\mathcal{L}_{\text{test}})$, where $\mathcal{L}_{\text{test}} = -\frac{1}{N}\sum_{i=1}^{N} \log p_\theta(y_i \mid x_i)$.

%-------------------------------------------------------------------
\section{Experimental Results}
\label{sec:results}
%-------------------------------------------------------------------

\subsection{Teacher Model Performance}

Table~\ref{tab:teacher-section} shows DeepSeek-R1 accuracy across competition sections. The teacher performs well in all three sections, supporting its use as a CoT data source.

\begin{table}[H]
  \caption{DeepSeek-R1 Teacher Accuracy by Section}
  \label{tab:teacher-section}
  \centering
  \begin{tabular}{lccc}
    \toprule
    \textbf{Section} & \textbf{Total} & \textbf{Correct} & \textbf{Accuracy} \\
    \midrule
    Freshman/Sophomore  & 265 & 239 & 90.2\% \\
    Junior/Senior       & 268 & 247 & 92.2\% \\
    Two-Person Speed    & 118 & 108 & 91.5\% \\
    \midrule
    Overall             & 651 & 594 & 91.4\% \\
    \bottomrule
  \end{tabular}
\end{table}

\subsection{Training Dynamics and Overfitting}

\begin{figure}[H]
  \centering
  \includegraphics[width=\textwidth]{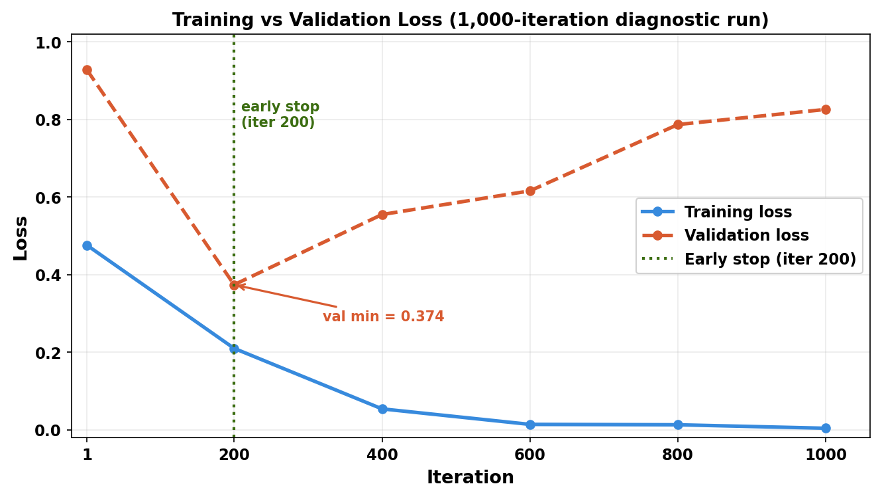}
  \caption{Training vs.\ validation loss over 1,000 iterations. Validation loss reaches its minimum (0.374) at iteration~200 (dotted line), indicating overfitting on a small corpus beyond this point.}
  \label{fig:training}
\end{figure}

Fig.~\ref{fig:training} shows the training and validation loss curves from the diagnostic 1,000-iteration run. Training loss fell from 0.476 at iteration 10 to near-zero by iteration 900. Validation loss followed a U-shaped trajectory: it reached a minimum of 0.374 at iteration 200, then rose steadily to 0.826 at iteration 1,000. Table~\ref{tab:loss} lists key checkpoints. This pattern is typical of small-corpus LoRA fine-tuning: with a few hundred unique training problems, the 11.5M trainable parameters are sufficient to memorize training examples within the first few hundred iterations. Up to iteration 200, the adapter weights capture generalizable reasoning patterns; thereafter, training shifts toward copying exact token sequences from the training set.

\begin{table}[H]
  \caption{Training and Validation Loss at Key Checkpoints}
  \label{tab:loss}
  \centering
  \begin{tabular}{cccc}
    \toprule
    \textbf{Iteration} & \textbf{Train Loss} & \textbf{Val Loss} & \textbf{Val PPL} \\
    \midrule
    1    & {---}  & 0.928 & {---}  \\
    200  & 0.210  & 0.374 & 1.454  \\
    400  & 0.054  & 0.555 & 1.742  \\
    600  & 0.014  & 0.616 & 1.852  \\
    800  & 0.013  & 0.787 & 2.197  \\
    1000 & 0.004  & 0.826 & 2.284  \\
    \bottomrule
  \end{tabular}
\end{table}

\subsection{Model Accuracy Comparison}

Table~\ref{tab:accuracy} summarizes accuracy across all model configurations and per-seed stability. The base Qwen2.5-7B achieves 64.67\%. The 1,000-iteration overfitted model reaches only 67.59\%, showing that continued training beyond iteration 200 does not improve answer accuracy. The five 200-iteration runs give a mean of 69.43\% $\pm$ 0.17\%; the narrow per-seed spread (0.17 pp) confirms the improvement is not an artifact of initialization. The two-person speed section is weakest at 64.3\%, consistent with its smaller share of training examples.

\begin{table}[H]
  \caption{Model Accuracy: Overall Comparison, Per-Seed Runs, and Section Breakdown}
  \label{tab:accuracy}
  \centering
  \begin{tabular}{lcc}
    \toprule
    \textbf{Configuration} & \textbf{Correct / Judged} & \textbf{Accuracy} \\
    \midrule
    \multicolumn{3}{l}{\textit{Overall model comparison}} \\
    DeepSeek-R1 (Teacher)         & 594 / 651 & 91.40\% \\
    Qwen2.5-7B Base               & 421 / 651 & 64.67\% \\
    Qwen2.5-7B (1k iter, overfit) & 440 / 651 & 67.59\% \\
    Qwen2.5-7B (200 iter, mean)   & --- / 651 & 69.43\% $\pm$ 0.17\% \\
    \midrule
    \multicolumn{3}{l}{\textit{Per-seed runs (200 iterations each)}} \\
    Seed 42   & 452 / 651 & 69.43\% \\
    Seed 123  & 451 / 651 & 69.28\% \\
    Seed 256  & 452 / 651 & 69.43\% \\
    Seed 512  & 453 / 651 & 69.58\% \\
    Seed 999  & 452 / 651 & 69.43\% \\
    \midrule
    \multicolumn{3}{l}{\textit{Section-level accuracy (200-iter, 5-run mean $\pm$ std)}} \\
    Freshman/Sophomore & --- & 70.0\% $\pm$ 0.19\% \\
    Junior/Senior      & --- & 71.0\% $\pm$ 0.16\% \\
    Two-Person Speed   & --- & 64.3\% $\pm$ 0.16\% \\
    \bottomrule
  \end{tabular}
\end{table}

\subsection{Test Loss and MATH-500 Generalization}

The 200-iteration fine-tuned models achieve mean test loss 1.258 and perplexity 3.519, compared to 1.837 and 6.280 for the overfitted model. The base Qwen2.5-7B achieves 70.1\% on MATH-500; the five fine-tuned models achieve 73.1\% $\pm$ 0.18\%, confirming that CoT reasoning patterns learned during distillation transfer to an out-of-domain problem set. Table~\ref{tab:math500-subject} shows the per-subject breakdown. The model is strongest on Algebra (93.5\%) and Number Theory (82.3\%), which are well represented in the John O'Bryan corpus, while Intermediate Algebra (52.1\%), Geometry (58.5\%), and Precalculus (58.9\%) remain weaker. These subject-level gaps suggest that future work should include more varied problem types in the training corpus to broaden generalization.

\begin{table}[H]
  \caption{Fine-Tuned Model on MATH-500 by Subject (200-Iter, 5-Run Mean $\pm$ Std)}
  \label{tab:math500-subject}
  \centering
  \begin{tabular}{lccc}
    \toprule
    \textbf{Subject} & \textbf{Total} & \textbf{Mean Acc.} & \textbf{Std Dev} \\
    \midrule
    Algebra                  & 124 & 93.5\% & $\pm$0.14\% \\
    Number Theory            &  62 & 82.3\% & $\pm$0.19\% \\
    Prealgebra               &  82 & 82.9\% & $\pm$0.17\% \\
    Counting \& Probability  &  38 & 60.5\% & $\pm$0.22\% \\
    Precalculus              &  56 & 58.9\% & $\pm$0.24\% \\
    Geometry                 &  41 & 58.5\% & $\pm$0.21\% \\
    Intermediate Algebra     &  96 & 52.1\% & $\pm$0.26\% \\
    \midrule
    Overall                  & 499 & 73.1\% & $\pm$0.18\% \\
    \bottomrule
  \end{tabular}
\end{table}

%-------------------------------------------------------------------
\subsection{Effect of Response Length on Accuracy}
\label{sec:token-budget}
%-------------------------------------------------------------------

Table~\ref{tab:token-combined} shows response length and overall accuracy for each reasoning level (R1--R6). R1 is the unrestricted baseline (mean 220 words, 69.43\%). Moving to R2 (400-token budget, mean 119.8 words) already costs 7.5 percentage points (61.9\%). Accuracy falls consistently to 41.9\% at R6 (100-token budget, mean 31.2 words). The drop from R2 to R3 is modest (1.3 points), but decline accelerates at R4 and below, where the model can no longer fit meaningful intermediate steps within the token budget.

\begin{table}[H]
  \caption{Response Length and Accuracy by Reasoning Level (R1--R6)}
  \label{tab:token-combined}
  \centering
  \begin{tabular}{lcccccc}
    \toprule
    \textbf{Level} & \textbf{Token Limit} & \textbf{Mean W} & \textbf{Med W}& \textbf{Correct} & \textbf{Total} & \textbf{Accuracy} \\
    \midrule
    R1 (base) & Unlimited & 220.0 & 218.0 & 452 & 651 & 69.43\% \\
    R2        & 400       & 119.8 & 105.6 & 370 & 598 & 61.9\%  \\
    R3        & 320       &  96.5 &  80.0 & 364 & 601 & 60.6\%  \\
    R4        & 180       &  55.6 &  45.0 & 307 & 601 & 51.1\%  \\
    R5        & 135       &  41.6 &  34.0 & 261 & 601 & 43.4\%  \\
    R6        & 100       &  31.2 &  27.0 & 252 & 601 & 41.9\%  \\
    \bottomrule
  \end{tabular}
\end{table}

Table~\ref{tab:token-section} breaks accuracy down by competition section. The Freshman/Sophomore and Junior/Senior sections show moderate declines from R2 to R6 (approximately 15 and 19 points respectively). The two-person speed section is far more sensitive: accuracy falls from 57.4\% at R2 to 22.9\% at R6, a drop of 34.5 points. This section features multi-step numerical problems where intermediate calculations are essential; a small token budget removes the space needed to carry those steps.

\begin{table}[H]
  \caption{Accuracy by Section and Reasoning Level}
  \label{tab:token-section}
  \centering
  \begin{tabular}{lccccc}
    \toprule
    \textbf{Section} & \textbf{R2} & \textbf{R3} & \textbf{R4} & \textbf{R5} & \textbf{R6} \\
    \midrule
    Freshman/Sophomore & 61.2\% & 63.8\% & 58.5\% & 44.7\% & 46.3\% \\
    Junior/Senior      & 64.5\% & 61.0\% & 52.4\% & 46.7\% & 45.9\% \\
    Two-Person Speed   & 57.4\% & 52.3\% & 31.2\% & 33.0\% & 22.9\% \\
    \bottomrule
  \end{tabular}
\end{table}

\begin{figure}[H]
  \centering
  \includegraphics[width=\textwidth]{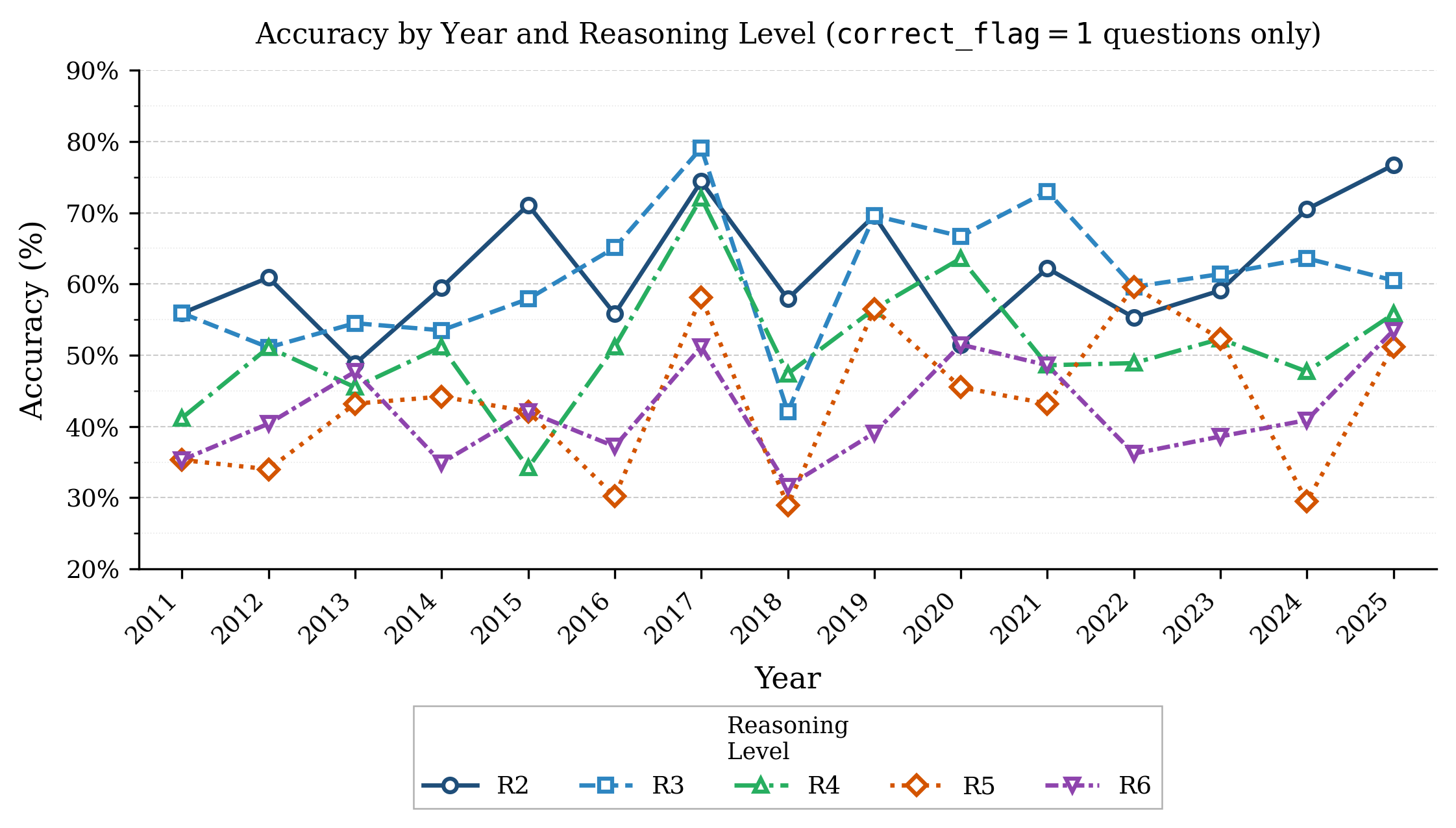}
  \caption{Accuracy by year and reasoning level (\texttt{correct\_flag == 1} questions only).}
  \label{fig:token-year}
\end{figure}

Fig.~\ref{fig:token-year} shows per-year accuracy across reasoning levels. Year 2017 stands out as consistently strong, ranging from 51.2\% (R6) to 74.4\% (R2), consistent with its lowest average difficulty score of 4.51 (Fig.~\ref{fig:difficulty}). Year 2018 (difficulty 5.79) collapses earlier under token constraints. Years 2013 and 2020 show relatively flat accuracy across levels, suggesting their problems can sometimes be answered with brief derivations.

\subsection{Architectural Compatibility and Inference}

Applying the Qwen2.5-7B adapters to a Qwen2.5-3B base raised a dimension mismatch (projection shape $3584\times8$ vs.\ hidden dim 2,048), confirming LoRA adapters are architecture-specific. On a representative prompt (interior-angle sum of a 19-gon), the fused model correctly computes $17\times180^\circ=3060^\circ$, generating 228 tokens at 31.8 tok/sec with 4.4 GB peak memory across all five variants.

%-------------------------------------------------------------------
\section{Ablation Study}
%-------------------------------------------------------------------

We conducted ablation experiments along three axes to isolate the contribution of each design choice: early stopping threshold, training seed stability, and token budget level.

\textbf{Early Stopping Threshold.} Models saved at iterations 200, 400, 600, 800, and 1,000 confirm that accuracy peaks at iteration 200 (69.43\%) and declines monotonically, reaching 67.59\% at iteration 1,000 (Table~\ref{tab:loss}). Test perplexity rises from 3.519 to 6.280 over the same range. Early stopping is therefore the decisive factor separating the generalizing model from the overfitted one.

\textbf{Seed Stability.} As shown in the per-seed block of Table~\ref{tab:accuracy}, five independent 200-iteration runs produced accuracy values with a standard deviation of 0.17 pp on the competition dataset and 0.18 pp on MATH-500. The narrow spread confirms the improvement is not an artifact of a particularly favorable initialization, and that a single well-timed run yields reliable results once the stopping point is established.

\textbf{Token Budget Levels.} Restricting generation length constitutes an implicit ablation on the CoT reasoning chain. Removing roughly half the words (R1 to R2, 220 to 119.8 words) costs 7.5 accuracy points. Cutting to approximately one quarter (R1 to R4, 220 to 55.6 words) costs 18.3 points. Below R4 the accuracy curve flattens, suggesting the model shifts to heuristic answer-guessing once the budget falls below approximately 50 words. The two-person speed section is most sensitive, losing 34.5 points from R2 to R6, while Freshman/Sophomore and Junior/Senior sections lose 15 and 19 points respectively.

\textbf{Error Type Distribution.} Of 197 incorrect responses examined qualitatively, 59.9\% (118 cases) involved genuine failures in the mathematical reasoning chain: hallucinated intermediate steps, incorrect formula application, or arithmetic errors. A further 32.0\% (63 cases) were extraction and formatting failures, where the model produced a correct mathematical result but expressed it in an unparseable form such as an unreduced fraction, an unrendered LaTeX string, or a sentence embedding the answer without a clear delimiter. The remaining 8.1\% (16 cases) were rounding or type mismatches. Taken together, approximately 40\% of all failures are attributable to output formatting rather than mathematical error, suggesting that a structured output parser or light post-processing step could recover a meaningful share of these cases without additional training.

%-------------------------------------------------------------------
\section{Discussion}
%-------------------------------------------------------------------

Our experiments yield two complementary findings. First, knowledge distillation with early stopping gives a reliable and transferable improvement: five independent runs produced consistent results (std dev 0.17 pp), and the 4.76 pp gain on the competition dataset is mirrored by a 3.0 pp gain on MATH-500. The subject-level MATH-500 results reveal that generalization is uneven: algebra and number theory, which are well represented in the John O'Bryan corpus, improve substantially, while geometry, precalculus, and intermediate algebra remain weak. This gap suggests that corpus composition has a direct influence on which reasoning skills transfer.

Second, response length has a large effect on accuracy. From R1 (220 words, 69.43\%) to R6 (31.2 words, 41.9\%), accuracy drops 27.5 points overall. The sharpest within-level decline occurs between R3 and R4 (96.5 to 55.6 words, 60.6\% to 51.1\%), identifying roughly 50--100 words as a practical lower bound for multi-step reasoning. The two-person speed section is most sensitive because its problems require carrying intermediate numerical results that cannot be compressed without loss; conceptual problems in the F/S and J/S sections are more robust to length reduction. The per-year difficulty analysis in Fig.~\ref{fig:difficulty} further shows that easier years sustain accuracy at shorter budgets while harder years collapse sooner, which aligns with the intuition that harder problems require proportionally more reasoning steps.

%-------------------------------------------------------------------
\section{Ethics and Responsibility}
%-------------------------------------------------------------------

The primary corpus comes from a single North American institution, which may bias the model toward Western pedagogical conventions and reasoning styles. Additionally, using DeepSeek-R1 as both teacher and verifier may introduce a stylistic dependency in the training data. At 69.43\% accuracy, this model is not ready for fully autonomous educational use. It is suitable as a drafting tool in human-in-the-loop setups, but all outputs should be reviewed before being shared with students. The token budget results add a practical deployment caution: imposing tight response length constraints substantially degrades accuracy on multi-step problems, and this degradation should be accounted for when configuring inference systems.

%-------------------------------------------------------------------
\section{Conclusion}
%-------------------------------------------------------------------

This paper described a knowledge distillation pipeline for mathematical reasoning using the John O'Bryan Mathematics Competition dataset. A DeepSeek-R1 dual-agent framework produces high-quality CoT training data (91.40\% verified accuracy), and LoRA fine-tuning with early stopping improves Qwen2.5-7B from 64.67\% to 69.43\% $\pm$ 0.17\%, with out-of-domain generalization confirmed on MATH-500 (73.1\% $\pm$ 0.18\% vs.\ 70.1\% base). The six-level token budget experiment shows that accuracy declines consistently from R1 (220 words, 69.43\%) to R6 (31.2 words, 41.9\%), with the steepest drop between R3 and R4, establishing a practical lower bound of roughly 50--100 words for multi-step competition-level reasoning. Five-seed stability analysis confirms that the improvement is robust to initialization (std dev 0.17 pp), and error type analysis reveals that approximately 40\% of failures stem from formatting rather than mathematical errors, pointing to post-processing as a tractable path to further gains. Future work will target larger and more diverse training corpora, section-stratified sampling, and explicit training on compressed CoT traces to help the model maintain accuracy under tighter token budgets.

%-------------------------------------------------------------------
\begin{credits}
\subsubsection{\ackname}
The authors thank the Department of Mathematics and Statistics at Northern Kentucky University for providing access to historical John O'Bryan Mathematics Competition materials.
\subsubsection{\discintname}
The authors have no competing interests to declare that are relevant to the content of this article.
\end{credits}

%-------------------------------------------------------------------

\end{document}